\RequirePackage{fix-cm}
\documentclass[twocolumn]{svjour3}          
\smartqed  
\usepackage{CJKutf8}

\usepackage{lmodern}
\usepackage{mdwlist}
\usepackage{url}
\usepackage[hidelinks,bookmarks=false]{hyperref}
\usepackage[official]{eurosym}
\usepackage{epsfig, times}
\usepackage{longtable}
\usepackage{fancyhdr}
\usepackage{algorithmic}
\usepackage{amsmath}

\usepackage{float}
\usepackage{verbatim}
\usepackage{afterpage}
\usepackage{tablefootnote}
\usepackage[toc,acronym,xindy]{glossaries}
\usepackage[colorinlistoftodos,prependcaption,textsize=tiny]{todonotes}

\usepackage{color}
\usepackage{amsfonts}
\usepackage{tabularx}
\usepackage[T1]{fontenc}
\usepackage[utf8]{inputenc}
\usepackage{graphicx}
\usepackage{adjustbox}
\usepackage{subfigure}
\usepackage{arabtex}
\usepackage{utf8}
\usepackage{gensymb}

\usepackage{epsfig}
\usepackage{wrapfig}
\usepackage[ruled]{algorithm2e}

\usepackage{multirow}
\usepackage{amssymb,amsmath,amsthm,enumitem}

\begin{document}
\setlength{\abovedisplayskip}{3pt}
\setlength{\belowdisplayskip}{3pt}
\title{Deriving Emotions and Sentiments from Visual Content: A Disaster Analysis Use Case}


\author{\mbox{Kashif Ahmad}         \and
        \mbox{Syed Zohaib}         \and
        \mbox{Nicola Conci}         \and  
	    \mbox{Ala Al-Fuqaha}       
       }


\institute{Hamad Bin Khalifa University \at
              \email{kahmad@hbku.edu.qa}           
}

\date{Received: date / Accepted: date}

\maketitle

\begin{abstract}

Sentiment analysis aims to extract and express a person's perception, opinions and emotions towards an entity, object, product and a service, enabling businesses to obtain feedback from the consumers. The increasing popularity of the social networks and users' tendency towards sharing their feelings, expressions and opinions in text, visual and audio content has opened new opportunities and challenges in sentiment analysis. While sentiment analysis of text streams has been widely explored in the literature, sentiment analysis of images and videos is relatively new. This article introduces visual sentiment analysis and contrasts it with textual sentiment analysis with emphasis on the opportunities and challenges in this nascent research area. We also propose a deep visual sentiment analyzer for disaster-related images as a use-case, covering different aspects of visual sentiment analysis starting from data collection, annotation, model selection, implementation and evaluations. We believe such rigorous analysis will provide a baseline for future research in the domain.

\keywords{Sentiment Analysis \and Analysis \and Emotions \and Deep Learning \and Multimedia Retrieval \and Natural Disasters}

\end{abstract}

\section{Introduction}
\section{Introduction}
Sentiment analysis, which aims to analyze and extract opinions, views and perceptions about an entity (e.g., product, service or an action), has been widely adopted by businesses helping them to understand consumers' perception about their products and services. The recent development and popularity of social media, in particular Twitter, Facebook and Instagram, helped researchers to extend the scope of sentiment analysis to other interesting applications. A recent example is reported by Ozturk et al. \cite{ozturk2018sentiment} where computational sentiment analysis is applied to the leading media sources as well as social media to extract sentiments on the Syrian's refugee crisis. Another example is reported by Kuvsen et al. \cite{kuvsen2018politics} where the neutrality of tweets and other reports from winner of the Austrian presidential election were analyzed and compared to the opponents' content on social media, which resulted in more negative sentiment scores.

The concept of sentiment analysis has been widely utilized in Natural Language processing (NLP), where several techniques have been employed to extract sentiments from text streams in terms of positive, negative and neutral perception/opinion. With the recent advancement in NLP techniques, an in-depth analysis of text streams from different sources is possible in different application domains, such as education, entertainment, hosteling and other businesses \cite{sadr2019robust}. More recently, several efforts have been made to analyze visual contents to derive sentiments. The vast majority of literature on visual sentiment analysis focuses on facial close-up images where facial expressions are used as visual cues to derive sentiments and predict emotions \cite{barrett2019emotional}. More recently, the concept of visual sentiment analysis has been extended to more complex images, including, for example, multiple objects and background details. The recent developments in machine learning and in particular deep learning have contributed to significantly boost the results also in this research area \cite{poria2018multimodal}. However, extracting sentiments from visual contents is not straightforward and several factors need to be considered.

In this article, we analyze the problem of visual sentiment analysis from different perspectives with particular focus on the challenges, opportunities, and potential applications. We also propose and present an automatic deep sentiment analyzer for disaster-related images as a use-case, and discuss the processing pipeline of visual sentiment analysis starting from data collection and annotation via a crowd-sourcing study, and conclude with the development and training of deep learning models. In fact, disaster-related images are complex and generally involve several objects as well as significant details in their backgrounds. We believe, such a challenging use-case is quintessential since it provides an opportunity to discuss the processing pipeline of visual sentiment analysis and provide a baseline for future research in the domain. Moreover, visual sentiment analysis of disasters, on its own right, has several applications and can contribute toward more livable communities. It can also help news agencies to cover such adverse events from different angles and perspectives. Similarly, humanitarian organizations can use such analysis to spread the information on a wider scale, focusing on the visual content that best demonstrates the evidence of a certain event.  

The  rest  of  the  paper  is  organized  as  follows:  Section  \ref{visual_sentiment} provides a general overview of visual sentiment analysis and contrasts it with textual sentiment analysis with emphasis on the opportunities, challenges and potential applications in this nascent research area. Section \ref{sentiment_analyzer} describes the proposed sentiment analyzer for natural disaster events; Section \ref{conclusion} concludes this study and provides directions of future research.

\section{Visual Emotion/Sentiment Analysis: Concepts and Background}
\label{visual_sentiment}


As implied by the popular proverb \textit{"a picture is worth a thousand words"}, visual contents are an effective mean to convey not only facts but also cues about sentiments and emotions. Such cues representing the emotions and sentiments of the photographers may trigger similar feelings at the observer and could be of help in understanding visual contents beyond semantic concepts in different application domains, such as education, entertainment, advertisement and journalism. Leveraging the power of visual contents, masters of photography have always utilized smart choices, especially in terms of scenes, perspective, angle of shooting and color filtering, to let the underlying information smoothly flow to the general public. In a similar way, every user aiming to increase in popularity on the Internet will utilize the same tricks. However, it is not fully clear how such emotional cues can be derived from visual contents and more importantly how the sentiments derived automatically from a scene by an automatic algorithm can be expressed. This opens an interesting line of research to interpret emotions and sentiments perceived by users viewing visual contents.   

In the literature, emotions, opinion mining, feelings and sentiment analysis have been used interchangeably \cite{munezero2014they,barrett2019emotional}. In practice, there is a significant difference among those terms. Sentiments are influenced by emotions and they allow individuals to show their emotions through expressions. In short, sentiments can be defined as a combination of emotions and cognition (i.e., sentiments are a combination of emotions and cognition). Therefore, Sentiments reveal underlying emotions through ways that require cognition (e.g., speech, actions, or written content).  

\begin{figure}[ht!]
\centering
\includegraphics[width=0.9\linewidth]{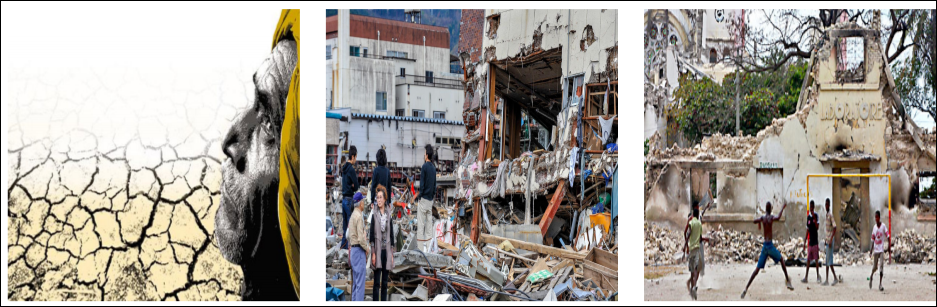}
\caption{Sample natural disaster-related images for sentiment analysis showing the diversity in the content and information to be extracted during sentiment analysis.}
	\label{fig:sample_images}
\end{figure}

The categorical representation of those concepts (i.e., emotions, sentiments) can be different, although the visual cues representing them are  closely related. For instance, the majority of efforts on emotion recognition, opinion mining and sentiment analysis express them in terms of three classes; namely, happy, sad and neutral or, similarly, positive, negative and neutral \cite{kim2018building}. However, similar types of visual features are used to infer those states \cite{soleymani2017survey}. For instance, facial expressions have been widely explored for both emotion recognition and visual sentiment analysis in close-up images \cite{barrett2019emotional}; though it would be simplistic to limit the capability of recognizing emotions and sentiments to face-close up images. There are several application domains where complex images with several objects and background details need to be analyzed. This is the case in the aforementioned scenario of disaster-related images, in which the background information is often crucial to evoke someone's emotions and sentiments. Figure \ref{fig:sample_images} provides samples of disaster-related images, highlighting the diversity in terms of content that needs to be examined. 

\subsection{Applications of Visual Sentiment Analysis}
\label{sec:aaplications}
Visual sentiment analysis is still in its infancy with a lot of potential to discover and unlock. The trend of an automatic analysis of visual contents for sentiments, opinion mining and emotion recognition has opened new directions of research, allowing the research community and the industry to explore new opportunities in exciting applications. Some of these applications include:

\begin{itemize}
   
    \item \textbf{Education}: Visual sentiment analyzers could be very productive in the education sector in various ways. For example, they could help to automatically filter and summarize visual educational contents. They could also help to perceive students' interest and learning experience. 
    
    \item \textbf{Entertainment}: wouldn't it be interesting to automatically extract exciting scenes from long video clips? Visual sentiment analysis could be instrumental to summarize TV programs and to extract affecting content, like sentiments, feelings, emotions and mood \cite{constantin2019computational}.  
    
    \item \textbf{News Broadcasting}: Visual sentiment analysis and emotion recognition could be exploited by the news agencies to set a particular tone or convey a specific message during news broadcasting, for a more focused information delivery. 
    
     \item \textbf{Fashion: Brand monitoring and advertising}: visual sentiment analysis can be very effective in the fashion industry and business intelligence, helping marketing analysts and companies with brand monitoring and grasping customers' preferences. 
     
     \item \textbf{A Tool for Humanitarian Organizations}: visual sentiment analysis could also have a social scope. For example, different humanitarian and no-profit organizations could benefit from such analysis to spread information for a good cause. 
     
\end{itemize}

\subsection{Key challenges in Visual Sentiment Analysis}
\label{sec:challenges}
In contrast to textual sentiment analysis, visual sentiment analysis is a nascent area of research, and several aspects still need to be investigated. The following are some of the key open research challenges that need to be addressed:

\begin{itemize}
     
     \item \textbf{Defining/identifying sentiments}: The biggest challenge in this domain is defining sentiments and identifying the one that better suits given visual content. Sentiments are very subjective and vary from person to person. For example, an image could be very destructive, shocking or appealing for one person and it may be just a neutral for another person. Moreover, the degree of the sentiments conveyed by an image in terms of its positivist or negativity is also another challenge to be tackled. 
    
    \item \textbf{Semantic gap}: one of the open questions that researchers have thoroughly investigated in the past decades is the semantic gap between the visual features and the cognition \cite{soleymani2017survey}. The selection of visual features is very crucial in multimedia analysis in general and in sentiment analysis in particular. We believe object and scene-level features could help in extracting such visual cues.
    
    \item \textbf{Expression of sentiments}: Are the three categories, \textit{Positive}, \textit{Negative} and \textit{Neutral}, enough to express/represent sentiments? What is the degree of sentiment positivist/negativity in a given image?, are other questions need to be answered; especially, considering the fact that sentiments are subjective and may vary from one person to another.   
    
   \item \textbf{Data collection and annotation}: image sources, sentiments labels, and feature selection are application-dependent, as an entertainment or education context is completely different from the humanitarian one. Such diversity makes it difficult to collect benchmark datasets from which knowledge can be transferred; thus, requiring ad-hoc data crawling and annotation. 
\end{itemize}

\section{Visual Sentiment Analysis of Disasters: a Use-case}
\label{sentiment_analyzer}

In this section, we propose a deep visual sentiment analyzer for disaster-related images as a use-case of visual sentiment analysis. Considering the complexity and the details available in disaster-related imagery, we believe visual sentiment analysis of disasters images could serve as a good use-case to explore the challenges and opportunities associated with visual sentiment analysis of complex images.

In the next subsections, we provide detailed description of the proposed methodology/pipeline, the crowd-sourcing study conducted for collecting ground-truth, experimental setup, conducted experiments, experimental results and lessons learned from the use-case. 

\subsection{Proposed Visual Sentiment Analysis Processing Pipeline}

Figure \ref{fig:methodology} provides the block diagram of the framework we propose for visual sentiment analysis. There are five main components of the proposed framework. Initially, social media platforms are crawled for disaster-related images followed by a data driven technique to select tags/sentiments, which can be associated with the disaster-related images, for the crowd-sourcing study. In the crowd-sourcing study, a subset of the downloaded images, after removing the irrelevant images, are annotated with human participants. A CNN and a transfer learning based method are then used for multi-label classification and to automatically assign sentiments/tags to the images. In the next subsections, we provide a detailed description of these components.

\begin{figure*}[ht!]
\centering
\includegraphics[width=0.9\linewidth]{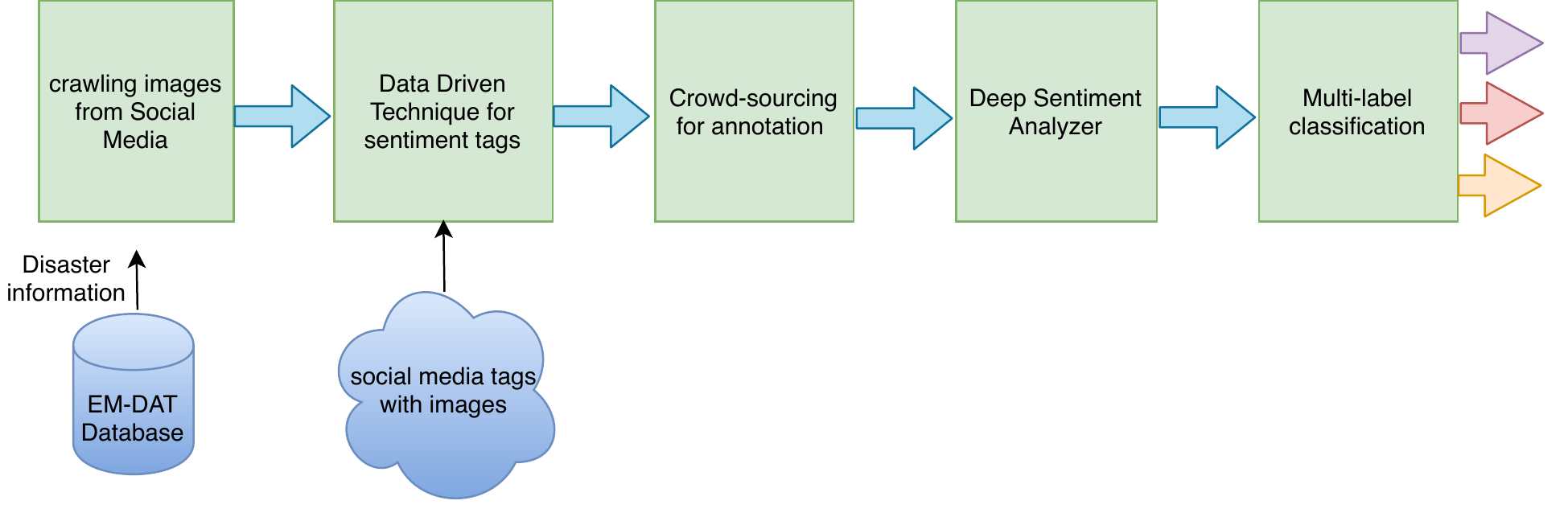}
\caption{Block diagram of the proposed visual sentiment analysis processing pipeline.}
	\label{correlation_tags}
\end{figure*}

\subsubsection{Crawling and sentiment category selection via a Data Driven Technique}
At the beginning of the processing pipeline, social media platforms are crawled to collect images for analysis. Images are downloaded using different keywords, such as floods, hurricanes, wildfires, droughts, landslides, earthquakes, etc. In order to obtain more relevant images, social media outlets are crawled with extended keywords, such as cyclones in Fiji, floods in Pakistan, etc. To this aim, the list of recent natural disasters is obtained from EM-DAT\footnote{https://www.emdat.be/}, which is a platform maintained by the United Nations providing statistics on world-wide disasters. 

The selection of labels for the crowd-sourcing study is one of the challenging and perhaps most crucial phases of the work. In the literature, sentiments are generally represented as \textit{Positive}, \textit{Negative} and \textit{Neutral} \cite{soleymani2017survey}. However, considering the nature and potential applications of the proposed deep sentiment analysis processing pipeline, we aim to target sentiments that are more specific to disaster-related contents. For instance, terms like pain, shock, and destruction are more commonly used with disaster-related contents. In order to choose more relevant and representative labels for our deep sentiment analyzer, we rely on a data driven technique where the tags and other textual details associated with images are analyzed to extract the relevant tags and tokens that describe such images.  

\subsubsection{The crowd-sourcing study}
\begin{figure*}[ht!]
\centering
\includegraphics[width=0.9\linewidth]{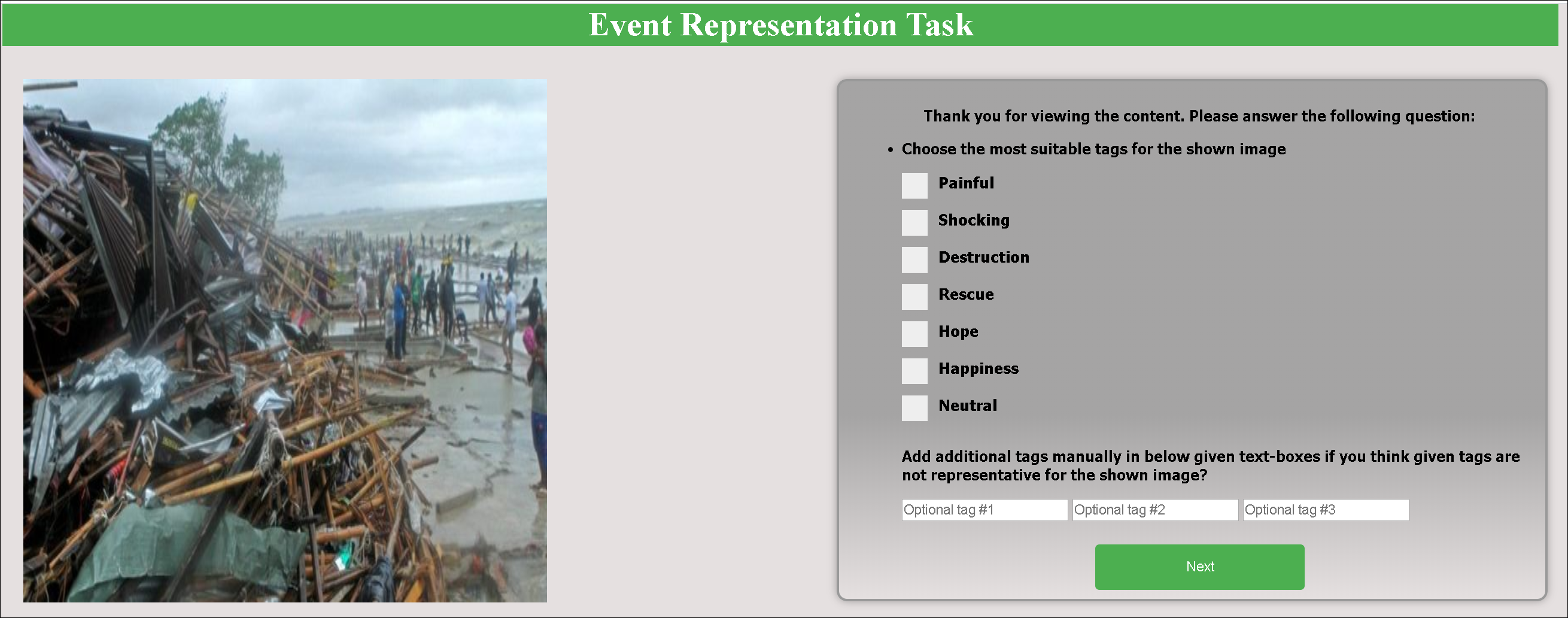}
\caption{Illustration of the web application used for the crowd-sourcing study. A disaster-related image and several tags are presented to the users for association. The users' are also encouraged to provide additional tags.}
	\label{crowdsourcing_study}
\end{figure*}

The crowd-sourcing study aims to develop ground-truth for the proposed deep sentiment analyzer by analyzing human's perceptions and sentiments about disasters and associated visual contents (i.e., images). The crowd-sourcing study is conducted through an online web application developed for this particular study where the selected images were presented to the participants to be annotated. The participants included individuals from different age groups, nationalities and with no scientific background, who were approached via social contacts. However, a significant portion of them were students from University of Trento, Italy and University of Engineering and Technology, Peshawar, Pakistan. Figure \ref{crowdsourcing_study} provides a screen-shot of the web application used for the crowd-sourcing study. As can be seen, the participants were provided with a disaster-related image, randomly selected from the pool of images, along with a set of tags to be associated with the image; namely, pain, shock, destruction, rescue, hope, happiness and neutral. Some of the tags, such as pain, shock, hope, happiness and neutral represent people's emotions; however, the additional tags, such as \textit{rescue} and \textit{destruction}, are closely related to adverse events and can be useful in several applications. For instance,  news agencies, humanitarian, and non-governmental organizations (NGOs) can use these tags for portraying a particular message or appealing for donations and other help during rescue operations. The participants were allowed to assign as many tags as they feel relevant to the underlying image. Moreover, the participants were also encouraged to assign additional tags to the images, in case they felt that the provided tags were not relevant to the image. These additional tags also help to take the participants' viewpoints into account. 

In the crowd-sourcing study, a total of 400 images from seven different types of disasters were analyzed. A total of 2,587 responses were obtained from the participants for these images each having at least five responses from five different participants of the study. We observed that during the study some of the tags were frequently used compared to others as shown in the aggregate statistics of the crowd-sourcing study provided in Table \ref{tab:statistics_cs}. These statistics give an insight about the sentiments associated with disasters and show how people perceive and react to such adverse events. Moreover, some of the tags are jointly used to describe an image. For instance, `destruction' has been jointly used with `pain' and `shock' and `rescue,' 138, 124 and 112 times, respectively. Similarly `pain' and `shock' have been jointly used 95 times. Figure \ref{correlation_tags} provides the statistics of frequent tags that have been jointly utilized by the participants in the study. We believe that such frequent multi-label analysis will help in refining the correlation and the gap/boundary between different sentiments.   
\begin{table}[]
\caption{Statistics of the crowd-sourcing study in terms of how frequently individual tags are associated with images.}
\label{tab:statistics_cs}
\begin{center}
\begin{tabular}{|c|c|}
\hline
\textbf{Sentiments/tags} & \textbf{Count} \\ \hline
Destruction & 871 \\ \hline
Happiness & 145 \\ \hline
Hope & 353 \\ \hline
Neutral & 214 \\ \hline
Pain & 454 \\ \hline
Rescue & 694 \\ \hline
Shock & 354 \\ \hline
\end{tabular}
\end{center}
\end{table}

\begin{figure*}[ht!]
\centering
\includegraphics[width=0.9\linewidth]{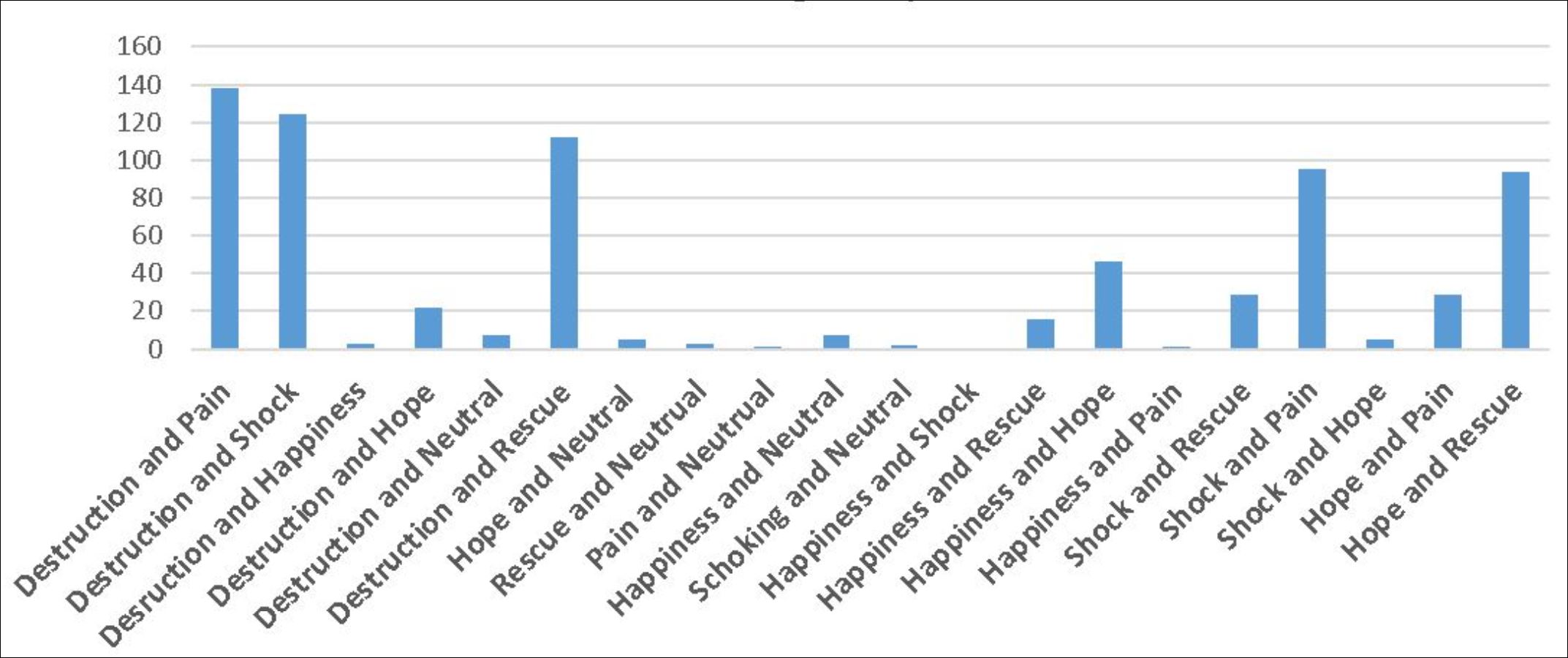}
\caption{Statistics of the crowd-sourcing study in terms of how frequently multiple tags are jointly associated with an image.}
	\label{correlation_tags}
\end{figure*}

\subsubsection{Deep Visual Sentiment Analyzer}
Our proposed multi-label deep visual sentiment analyzer is mainly based on a CNN and transfer learning. Based on our previous experience and considering the nature of the application, we rely on both object and scene level features extracted through existing deep models pre-trained on the ImageNet \cite{deng2009imagenet} and Places \cite{zhou2014learningB} datasets. The model pre-trained on ImageNet extract object-level information while the ones pre-trained on the Places dataset cover the background details. To this aim, we employed four different state-of-the-art deep models; namely, AlexNet \cite{krizhevsky2012imagenet}, VggNet \cite{simonyan2014very}, ResNet \cite{he2016deep} and Inception v-3 \cite{szegedy2016rethinking}. These models are fine-tuned on the newly collected dataset for visual sentiment analysis of  disaster-related images. 

The object and scene level features are also combined using early fusion by including a concatenation layer in our framework, where features from models pre-trained on the ImageNet and Places datasets are combined before the classification layer. In the current implementation, we rely on simple fusion technique aiming to identify and analyze the potential improvement by combining both object and background details for visual sentiment analysis of disaster-related images.

For the multi-label analysis, we made several changes in the framework to adapt the pre-trained models for the task. As a first step, a vector of the ground truth having all the possible labels has been created with the corresponding changes in the models. For instance, the top layer of the model has been modified to support multi-label classification by replacing the soft-max function with a sigmoid function. The sigmoid function helps our goal by presenting the results for each label in probabilistic terms while the soft-max function holds the probability law and squashes all the values of a vector into a  \textit{[0,1]} range. Similar changes (i.e., replacing softmax with sigmoid function) are made in the formulation of the cross entropy to properly fine-tune the pre-trained models. 

\subsection{Experiments and Evaluations}
The main motivation behind our experiments is to provide a baseline for future work in the domain. To this aim, we evaluate the proposed multi-label framework for visual sentiment analysis using several existing deep models pre-trained on the ImageNet and Places datasets. Table \ref{tab:results_60} provides the experimental results of our proposed deep sentiment analyzer. Over all, encouraging results are obtained with all deep models with slight improvement for Inception V-3 over the other models. Another aspect we want to analyze in the experiments is to evaluate the impact/effectiveness of object and scene level features in disaster analysis. As can be seen, a 0.5\% improvement has been observed in the accuracy of the VggNet model when trained on ImageNet (i.e., object-level information) over the Places dataset (i.e., scene-level information).  

As can be seen, accuracy of the framework has been improved by jointly utilizing models pre-trained on the ImageNet and Places datasets. The demonstrates the correctness of our hypothesis that object and background details can be jointly utilized to better represent the sentiments in visual contents. 


\begin{table}[]
\caption{Evaluation of the proposed visual sentiment analyzer with different deep learning models pre-trained on the ImageNet and Places datasets.}
\label{tab:results_60}
\begin{center}
\begin{tabular}{|c|c|}
\hline
\textbf{Model} & \textbf{Accuracy (\%)} \\ \hline
AlexNet (ImageNet) & 79.69 \\ \hline
VggNet (ImageNet) & 79.58  \\ \hline
 Inception-v3 (ImageNet) & 80.70  \\ \hline
ResNet (ImageNet) & 78.01 \\ \hline
VggNet (Places)   & 79.08 \\ \hline
 AlexNet  (Places)  & 79.61 \\ \hline
  Fusion (VggNet places + VggNet ImageNet)   & 81.34 \\ \hline
\end{tabular}
\end{center}
\end{table}

\subsection{Lessons learned}
The lessons learned from this use-case are summarized as follow:
\begin{itemize}
    \item Though the preliminary results of the deep visual sentiment analyzer are encouraging, visual sentiment analysis of disaster-related images is a very challenging task that required more investigation. More sophisticated frameworks with rich visual features could further improve the results.
    
    \item Sentiment analysis aims to extract people's perceptions of the images; thus, crowd-sourcing seems a suitable option for collecting training and ground truth datasets. However, choosing labels/tags for conducting a successful crowd-sourcing study is very challenging. A data driven approach can be more suitable compared to using pre-defined general sentiment categories.
    
    \item  Object and scene level (i.e., background) features are crucial in the visual sentiment analysis of disaster-related images. Their joint use better represents people's emotions and sentiments.
    
    \item  The capability of automatic visual sentiment analysis of disasters-related images will lead to better understanding of disaster events and will benefit a broad range of applications.
    
\end{itemize}

\section{Summary: Conclusion and Future research directions}
\label{conclusion}
In this article, we focused on the emerging concept of visual sentiment analysis. We mainly discussed the concepts and challenges associated with visual sentiment analysis along with the potential application domains which can benefit from it. We also focused on visual sentiment analysis of images from natural disasters as a use-case. As discussed in the article, visual sentiment analysis is an exciting research domain that will benefit users and the community in a diversified set of applications, such as entertainment, news casting, education and fashion. The current literature shows a tendency towards visual sentiment analysis of general images shared on social media by deploying deep learning techniques to extract object and facial expression based visual cues. However, we believe, as also demonstrated by the use-case presented in this study, visual sentiment analysis can benefit from the joint use of object and scene level information.

As described in Section \ref{sec:challenges}, there are several open research challenges in this nascent research area. One of the main challenges is defining the sentiment itself and the expression of sentiments. We believe this can be an interesting direction of future work to explore the link between vision and human psyche. Another interesting direction is to investigate data driven approaches that rely on the photographers' emotions and how they perceive the image that they capture. Another aspect of sentiment analysis, which is not fully explored yet, is the effective gap between visual cues/features and the cognition. We believe, there's a lot to be explored yet in this direction. This research area lacks a large-scale benchmark dataset covering multi-modal information for vertical application domains. There are some datasets mostly containing generic images with facial expressions. However, they are meant for basic sentiment analysis. We believe datasets applicable for visual sentiment analysis in vertical application domains can be helpful to introduce new applications and services.


\bibliographystyle{spmpsci}      

\bibliography{sigproc}

\end{document}